\documentclass[sigconf]{acmart}

\usepackage{amsmath}
\usepackage{bm}
\usepackage{mathrsfs}
\usepackage{amsthm}
\usepackage{amsfonts}
\usepackage{subfigure}
\usepackage{multirow}
\usepackage{url}
\usepackage{colortbl}
\usepackage[ruled,linesnumbered,vlined]{algorithm2e}
\usepackage{makecell}
\usepackage{enumitem}
\usepackage{diagbox}
\usepackage{float}

\AtBeginDocument{%
  \providecommand\BibTeX{{%
    \normalfont B\kern-0.5em{\scshape i\kern-0.25em b}\kern-0.8em\TeX}}}

\copyrightyear{2023}
\acmYear{2023}
\setcopyright{acmlicensed}\acmConference[WWW '23]{Proceedings of the ACM Web Conference 2023}{April 30-May 4, 2023}{Austin, TX, USA}
\acmBooktitle{Proceedings of the ACM Web Conference 2023 (WWW '23), April 30-May 4, 2023, Austin, TX, USA}
\acmPrice{15.00}
\acmDOI{10.1145/3543507.3583453}
\acmISBN{978-1-4503-9416-1/23/04}

\begin{document}

%%commands
\newcommand{\framework}{SE-GSL}
\newcommand\negimp[1]{\textcolor[RGB]{255,180,0}{#1}}
\newcommand\posimp[1]{\textcolor[RGB]{0,190,0}{#1}}
\theoremstyle{definition}
\newtheorem{define}{Definition}[]
\title{\framework: A General and Effective Graph Structure Learning Framework through Structural Entropy Optimization}

% \author{
% Dongcheng Zou$^1$,
% Xiang Huang$^1$,
% Hao Peng$^1$,
% Renyu Yang$^1$,
% Jia Wu$^2$,
% Jianxin Li$^1$,
% Chunyang Liu$^1$, and 
% Philip S. Yu$^3$
% }
% \affiliation{
% \institute{
% $^1$Beihang University;
% $^2$Macquarie University;
% $^3$University of Illinois Chicago
% }
% }
% \email{
% {zoudongcheng, huang.xiang, penghao, renyu.yang, lijx}@buaa.edu.cn, 
% jia.wu@mq.edu.au, 
% liuchunyang@didiglobal.com, 
% psyu@uic.edu
% }

\author{Dongcheng Zou}
\affiliation{%
  \institution{Beihang University}
  \city{Beijing}
  \country{China}
}
\email{zoudongcheng@buaa.edu.cn}

\author{Hao Peng}
\affiliation{%
  \institution{Beihang University}
  \city{Beijing}
  \country{China}
}
\email{penghao@buaa.edu.cn}
\authornote{Corresponding author}

\author{Xiang Huang}
\affiliation{%
  \institution{Beihang University}
  \city{Beijing}
  \country{China}
}
\email{huang.xiang@buaa.edu.cn}

\author{Renyu Yang}
\affiliation{%
  \institution{Beihang University}
  \city{Beijing}
  \country{China}
}
\email{renyu.yang@buaa.edu.cn}

\author{Jianxin Li}
\affiliation{%
  \institution{Beihang University}
  \city{Beijing}
  \country{China}
}
\email{lijx@buaa.edu.cn}

\author{Jia Wu}
\affiliation{%
  \institution{Macquarie University}
  \city{Sydney}
  %\state{NSW}
  \country{Australia}
}
\email{jia.wu@mq.edu.au}

\author{Chunyang Liu}
\affiliation{%
  \institution{Didi Chuxing}
  \city{Beijing}
  \country{China}
}
\email{liuchunyang@didiglobal.com}

\author{Philip S. Yu}
\affiliation{%
  \institution{University of Illinois Chicago}
  \city{Chicago}
  %\state{IL}
  \country{USA}
}
\email{psyu@uic.edu}

\renewcommand{\shortauthors}{Zou and Peng, et al.}

%%
%% The abstract is a short summary of the work to be presented in the article.
\begin{abstract}
Graph Neural Networks (GNNs) are de facto solutions to structural data learning. However, it is susceptible to low-quality and unreliable structure, which has been a norm rather than an exception in real-world graphs. Existing graph structure learning (GSL) frameworks still lack robustness and interpretability.
% in noisy or heterophily graphs. 
This paper proposes a general GSL framework, \framework{}, through structural entropy and the graph hierarchy abstracted in the encoding tree. Particularly, we exploit the one-dimensional structural entropy to maximize embedded information content when auxiliary neighbourhood attributes is fused to enhance the original graph. 
A new scheme of constructing optimal encoding trees is proposed to minimize the uncertainty and noises in the graph whilst assuring proper community partition in hierarchical abstraction. 
We present a novel sample-based mechanism for restoring the graph structure via node structural entropy distribution. It increases the connectivity among nodes with larger uncertainty in lower-level communities.
\ \framework\ is compatible with various GNN models and enhances the robustness towards noisy and heterophily structures. 
Extensive experiments show significant improvements in the effectiveness and robustness of structure learning and node representation learning.
\end{abstract}





\maketitle

\section{Introduction}
\label{sec:intro}

Graph Neural Networks (GNNs)~\cite{wu2020comprehensive,zhou2020graph} have become the cornerstone and de facto solution of structural representation learning. 
Most of the state-of-the-art GNN models employ message passing~\cite{gilmer2017neural} and recursive information aggregation from local neighborhoods ~\cite{velivckovic2017graph,leskovec2019powerful,peng2021reinforced,yang2023wsdm} to learn node representation. 
These models have been advancing a variety of tasks, including node classification~\cite{welling2016semi,xu2018powerful,peng2019event}, node clustering~\cite{bianchi2020spectral,peng2022event}, graph classification~\cite{ying2018hierarchical,Peng2020motif}, and graph generation~\cite{you2018graphrnn}, etc. 

GNNs are extremely sensitive to the quality of given graphs and thus require resilient and high-quality graph structures. 
However, it is increasingly difficult to meet such a requirement in real-world graphs. 
Their structures tend to be noisy, incomplete, adversarial, and heterophily (i.e., the edges with a higher tendency to connect nodes of different types), which can drastically weaken the representation capability of GNNs~\cite{pei2019geom,luo2021learning,dai2021nrgnn}.
Recent studies also reveal that even a minor perturbation in the graph structure can lead to inferior prediction quality ~\cite{bojchevski2019certifiable,zhang2020gnnguard,sun2022adversarial}. 
Additionally, GNNs are vulnerable to attacks since the raw graph topology is decoupled from node features, and attackers can easily fabricate links between entirely different nodes~\cite{zhang2020gnnguard,sun2022adversarial}.

To this end, Graph Structure Learning (GSL)~\cite{chen2020iterative,jin2020graph,wang2020gcn,zhu2021deep,zhu2021contrastive,SunGLS2022,Sun2022Position} becomes the recent driving force for learning superior task-relevant graph topology and for enhancing the resilience and robustness of node representation. 
% The existing works focus on parameterizing the adjacency matrix and on jointly optimizing GNN whilst imposing regularization constraints on refined graph structures ($e.g.$, sparsity, smoothness, and low-rank~\cite{chen2020iterative,luo2021learning,yang2023wsdm}).
The existing works focus on jointly optimizing GNN whilst imposing regularization on refined graph structures.
Typical methods include metric-based~\cite{chen2020iterative,wang2020gcn,li2018adaptive}, probabilistic sampling ~\cite{franceschi2019learning,rong2019dropedge,zheng2020robust}, and learnable structure approach~\cite{jin2020graph}, etc.
While promising, GNNs and GSL still have the following issues. 
\emph{i) robustness to system noises and heterophily graphs.} While many GSL models strive to fuse node features and topological features through edge reconstruction (e.g., add, prune, or reweight) ~\cite{wang2020gcn,zhang2020gnnguard,zhu2021contrastive}, additional noises and disassortative connections will be inevitably involved in the fused structure due to the unreliable priori topology and node embeddings, which would further degrade the GNNs representation capability~\cite{li2022reliable}. 
\emph{ii) model interpretability.}
Fully parameterizing the adjacency matrix will incur a non-negligible cost of parameter storage and updating and is liable to low model interpretability~\cite{gilpin2018explaining}. 
Although some studies on the improved GNN interpretability ~\cite{ying2019gnnexplainer,huang2022graphlime}, few works can effectively explain the topology evolution during graph structure learning.
Therefore, fusing the node and topological features in a noisy system environment to obtain GNN-friendly graphs by exploiting inherent graph structures is still an underexplored problem~\cite{wu2020graph}.

\begin{figure}[t]
  \centering
  \includegraphics[width=0.44\textwidth]{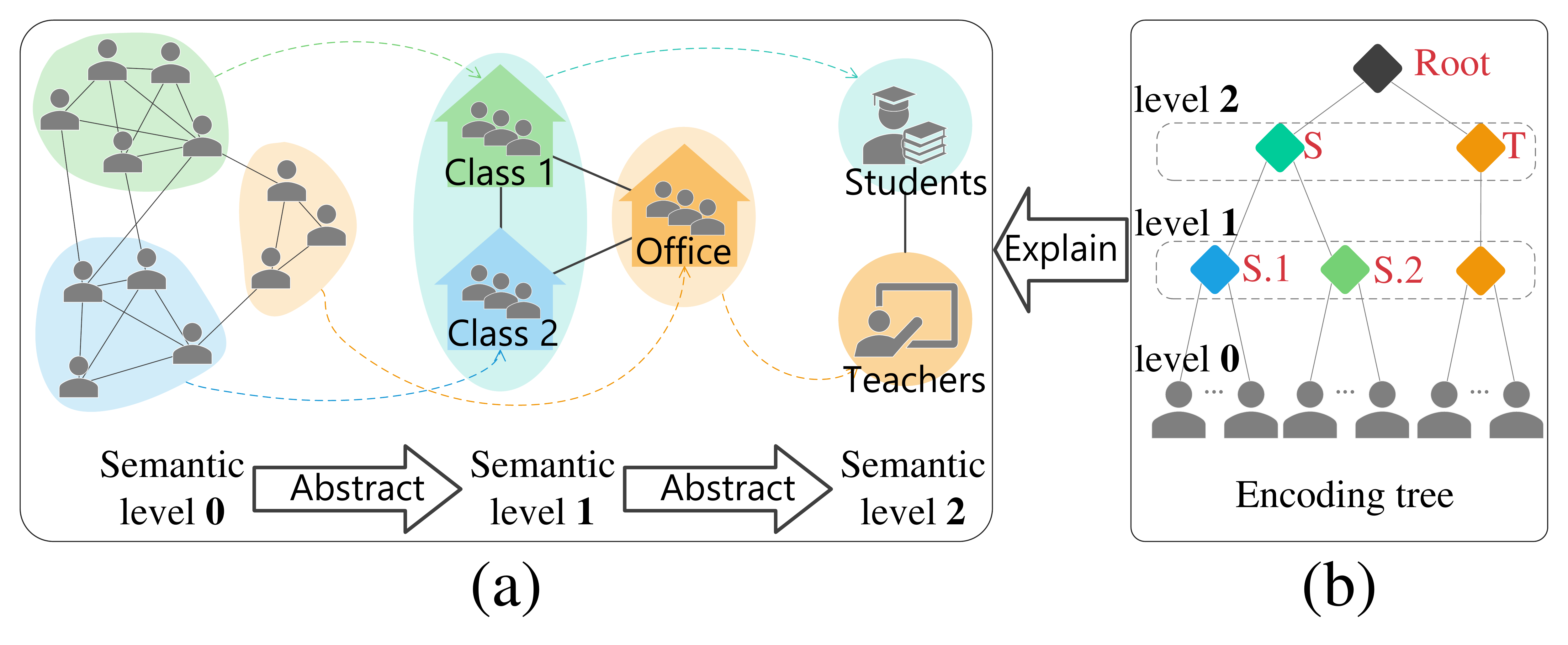}
  \caption {An illustrative example of the hierarchical community (semantics) in a simple social network. (1) Vertices and edges represent the people and their interconnectivity (e.g., common locations, interests, occupations). There are different abstraction levels, and each community can be divided into sub-communities in a finer-grained manner (e.g., students are placed in different classrooms while teachers are allocated different offices). The lowest abstraction will come down to the individuals with own attributes, and the highest abstraction is the social network system. (b) An encoding tree is a natural form to represent and interpret such a multi-level hierarchy.}
  % \caption {An illustrative example of the hierarchical community (semantics) in a simple social network. (1) Vertices and edges represent the people and their interconnectivity. Each community can be divided into sub-communities in a finer-grained manner at different abstraction levels (e.g., students are placed in different classrooms while teachers are allocated different offices). Each level of abstraction has realistic semantics (b) An encoding tree is a natural form to interpret such a multi-level hierarchy.
  % }
  \label{fig:intro}  
  \Description{An example of hierarchical communities in a simple social network.}
\end{figure}

In this paper, we present\ \framework, a general and effective graph structure learning framework that can adaptively optimize the topological graph structure in a learning-free manner and can achieve superior node representations, widely applicable to the mainstream GNN models. 
This study is among the first attempts to marry the structural entropy and encoding tree theory~\cite{li2016structural} with GSL, which offers an effective measure of the information embedded in an arbitrary graph and structural diversity. 
The multi-level semantics of a graph can be abstracted and characterized through an encoding tree.
Encoding tree~\cite{li2016structural,li2018decoding,zeng2023effective} represents a multi-grained division of graphs into hierarchical communities and sub-communities, thus providing a pathway to better interpretability.  
Fig.~\ref{fig:intro} showcases how such graph semantics are hierarchically abstracted. 
Specifically, we first enhance the original graph topology by incorporating the vertex similarities and auxiliary neighborhood information via the $k$-Nearest Neighbors ($k$-NN) approach, so that noise can be better identified and diminished. 
% In doing so, a $k$-Nearest Neighbors ($k$-NN) graph is generated and fused into the original graph structure. 
This procedure is guided by the $k$-selector that maximizes the amount of embedded information in the graph structure.
% We then propose a scheme to establish an optimal encoding tree by high-dimensional structural entropy minimization, with the intuition of minimizing the graph uncertainty and edge noises whilst maximizing the knowledge embedded in the hierarchical abstraction.
We then propose a scheme to establish an optimal encoding tree to minimize the graph uncertainty and edge noises whilst maximizing the knowledge in the encoding tree.
% The intuition is to minimize the uncertainty and edge noises in the graph whilst maximizing the knowledge embedded (e.g., optimal community partition) in the hierarchical abstraction. 
To restore the entire graph structure that can be further fed into GNN encoders, we recover edge connectivity between related vertices from the encoding tree taking into account the structural entropy distribution among vertices. 
The core idea is to weaken the association between vertices in high-level communities whilst establishing dense and extensive connections between vertices in low-level communities. The steps above will be iteratively conducted to co-optimize both graph structure and node embedding learning.
\framework\footnote{code is available at: https://github.com/RingBDStack/SE-GSL} is an interpretable GSL framework that effectively exploits the substantive structure of the graph. 
We conduct extensive experiments and demonstrate significant and consistent improvements in the effectiveness of node representation learning and the robustness of edge perturbations.
% We conduct extensive experiments on nine node classification datasets and demonstrate significant and consistent improvements in the effectiveness in node representation learning. 
% We artificially inject noisy edges into three typical datasets to evaluate the robustness of GSL in the event of graph perturbations.   

\textbf{Contribution highlights:}
i) \framework{} provides a generic GSL solution to improve both the effectiveness and robustness of the mainstream GNN approaches.
ii) \framework{} offers a new perspective of navigating the complexity of attribute noise and edge noise by leveraging structural entropy as an effective measure and encoding tree as the graph hierarchical abstraction. 
iii) \framework{} presents a series of optimizations on the encoding tree and graph reconstruction that can not only explicitly interpret the graph hierarchical meanings but also reduce the negative impact of unreliable fusion of node features and structure topology on the performance of GNNs. 
iv) We present a visualization study to reveal improved interpretability when the graph structure is evolutionary.

\section{Preliminaries}\label{sec:prelim}
This section formally reviews the basic concepts of Graph, Graph Neural Networks (GNNs), Graph Structure Learning (GSL), and Structural Entropy. 
Important notations are given in Appendix ~\ref{appendix:notations}.

\subsection{Graph and Graph Structure Learning}
\noindent \textbf{Graph and Community}. 
% \subsubsection{Graph and Communities}
Let $G = \{V,E,X\}$ denote a graph, where $V$ is the set of $n$ vertices\footnote{A vertex is defined in the graph and a node in the tree.}, $E \subseteq V \times V$ is the edge set, and $X \in \mathbb{R}^{n\times d}$ refers to the vertex attribute set.
$\mathrm {A} \in \mathbb{R}^{n \times n}$ denotes the adjacency matrix of $G$, where $\mathrm {A}_{ij}$ is referred to as the weight of the edge between vertex $i$ and vertex $j$ in $G$. 
Particularly, if $G$ is unweighted, $\mathrm {A} \in \{0,1\}^{n \times n}$ and $\mathrm {A}_{ij}$ only indicate the existence of the edges. 
In our work, we only consider the undirected graph, where $\mathrm {A}_{ij} = \mathrm {A}_{ji}$. 
For any vertex $v_i$, the degree of $v_i$ is defined as $d(v_i) = \sum_{j}\mathrm {A}_{ij}$, and $D = \mathrm {diag}(d(v_1),d(v_2),\dots,d(v_n))$ refers to the degree matrix.

Suppose that $\mathcal{P} =
\{P_1, P_2,\dots, P_L\}$ is a partition of $V$. 
Each $P_i$ is called a \textit{community} (aka. module or cluster), representing a group of vertices with commonality.
% such that the density of edges between vertices is higher than the density with the rest of the graph. 
Due to the grouping nature of a real-world network, each community of the graph can be hierarchically split into multi-level \textit{sub-communities}. 
Such \textit{hierarchical community} partition (i.e., hierarchical \textit{semantic}) of a graph can be intrinsically abstracted as the encoding tree~\cite{li2016structural,li2018decoding}, and each tree node represents a specific community. 
Take Fig.~\ref{fig:intro} as an example: at a high abstraction (semantic) level, the entire graph can be categorized as two coarse-grained communities, i.e., teachers (T) and students (S). 
Students can be identified as sub-communities like S.1 and S.2, as per the class placement scheme.

\noindent \textbf{Graph Structure Learning (GSL)}. 
For any given graph $G$, the goal of GSL~\cite{zhu2021deep} is to simultaneously learn an optimal graph structure $G^*$ optimized for a specific downstream task and the corresponding graph representation $Z$. 
In general, the objective of GSL can be summarized as $\mathcal{L}_{gsl} = \mathcal{L}_{task}(Z,Y) + \alpha \mathcal{L}_{reg}(Z,G^*,G)$,
% \begin{equation}\label{eq:LGSL}
% \end{equation}
where $\mathcal{L}_{task}$ refers to a task-specific objective with respect to the learned representation $Z$ and the ground truth $Y$.
$\mathcal{L}_{reg}$ imposes constraints on the learned graph structure and representations, and $\alpha$ is a hyper-parameter.

\subsection{Structural Entropy}
Different from information entropy (aka. Shannon entropy) that measures the uncertainty of probability distribution in information theory ~\cite{shannon1948mathematical}, \textit{structural entropy}~\cite{li2016structural} measures the structural system diversity, e.g., the uncertainty embedded in a graph.

\noindent \textbf{Encoding Tree}. 
Formally, the encoding tree $\mathcal{T}$ of graph $G=(V, E)$ holds the following properties:
\textbf{(1)} The root node $\lambda$ in $\mathcal{T}$ has a label $T_\lambda = V$, $V$ represents the set of all vertices in $G$.
\textbf{(2)} Each non-root node $\alpha$ has a label $T_\alpha \subset V$. 
Furthermore, if $\alpha$ is a leaf node, $T_\alpha$ is a singleton with one vertex in $V$.
\textbf{(3)} For each non-root node $\alpha$, its parent node in $T$ is denoted as $\alpha^-$.
\textbf{(4)} For each non-leaf node $\alpha$, its $i$-th children node is denoted as $\alpha^{\left \langle i \right \rangle }$ ordered from left to right as $i$ increases.
\textbf{(5)} For each non-leaf node $\alpha$, assuming the number of children $\alpha$ is $N$, all vertex subset $T_{\alpha^{\left \langle i \right \rangle}}$ form a partition of $T_\alpha$, written as $T_\alpha = {\textstyle \bigcup_{i=1}^{N}} T_{\alpha^{\left \langle i \right \rangle}}$ and ${\textstyle \bigcap_{i=1}^{N}} T_{\alpha^{\left \langle i \right \rangle}} = \varnothing$.
If the encoding tree's height is restricted to $K$, we call it \textit{$K$-level} encoding tree. 
Entropy measures can be conducted on different encoding trees.

\noindent \textbf{One-dimensional Structural Entropy}. \label{prelim:1dse}
In a single-level encoding tree $\mathcal{T}$, its structural entropy degenerates to the unstructured Shannon entropy, which is formulated as:
\begin{equation}\label{eq:H1}
H^1(G) = -\sum_{v \in V}{\frac{d_v}{vol(G)}\log_{2}{\frac{d_v}{vol(G)}}},
\end{equation}
where $d_v$ is the degree of vertex $v$, and $vol(G)$ is the sum of the degrees of all vertices in $G$.
According to the fundamental research~\cite{li2016structural}, one-dimensional structural entropy $H^1(G)$ measures the uncertainty of vertex set $V$ in $G$, which is also the upper bound on the amount of information embedded in $G$.

\noindent \textbf{High-dimensional Structural Entropy}. 
For the encoding tree $\mathcal{T}$, we define high-dimensional structural entropy of $G$ as:
\begin{equation}\label{eq:HK}
H^K(G) = \min_{\forall \mathcal{T}:height(\mathcal{T}) \le K}\{H^{\mathcal{T}}(G)\},
\end{equation}
\begin{equation}\label{eq:HT}
% \vspace{-0.1em}
H^{\mathcal{T}}(G) = \sum_{\alpha \in \mathcal{T},\alpha \ne \lambda} {H^{\mathcal{T}}(G;\alpha)} = -\sum_{\alpha \in \mathcal{T},\alpha \ne \lambda} {\frac{g_\alpha}{vol(G)}\log_{2}{\frac{\mathcal{V}_{\alpha}}{\mathcal{V}_{\alpha^-}}}},
\end{equation}
where $g_\alpha$ is the sum weights of the cut edge set $[T_\alpha,T_\alpha/T_\lambda]$, i.e., all edges connecting vertices inside $T_\alpha$ with vertices outside $T_\alpha$. $\mathcal{V}_\alpha$ is the sum of degrees of all vertices in $T_\alpha$. 
$H^{\mathcal{T}}(G;\alpha)$ is the structural entropy of node $\alpha$ and $H^{\mathcal{T}}(G)$ is the structural entropy of $\mathcal{T}$. 
$H^K(G)$ is the $K$-dimensional structural entropy, with the optimal encoding tree of $K$-level .%% 2-related work
\section{Our Approach}\label{sec:framework}

\begin{figure*}[t]
  \centering
  \includegraphics[width=0.98\textwidth]{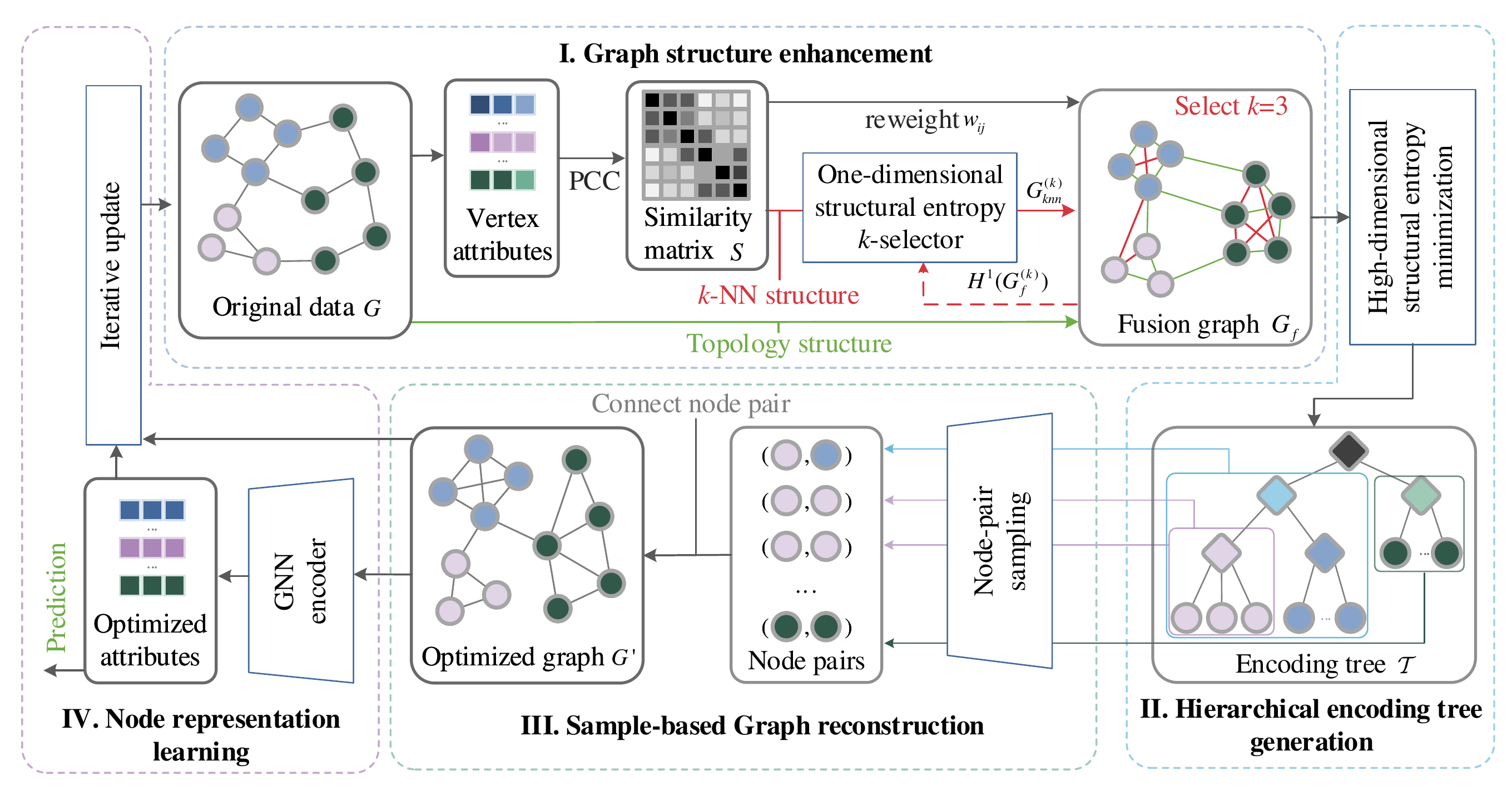}
  \caption{The overall architecture of\ \framework.}
%   \FIXME{change $G^{(k)}$ to $G_{knn}^{(k)}$}
  \label{fig:framework}
  \Description{The overall architecture of SE-GSL.}
\end{figure*}

This section presents the architecture of \framework{}, then elaborate on how we enhance the graph structure learning by structural entropy-based optimization of the hierarchical encoding tree.

\subsection{Overview of ~\framework}
Fig.~\ref{fig:framework} depicts the overall pipeline. 
At the core of ~\framework{} is the structure optimization procedure that transforms and enhances the graph structure. 
More specifically, it encompasses multi-stages: graph structure enhancement, hierarchical encoding tree generation, and sampling-based structure reconstruction before an iterative representation optimization.

First, the original topological information is integrated with vertex attributes and the neighborhood in close proximity. 
Specifically, we devise a similarity-based edge reweighting mechanism and incorporate $k$-NN graph structuralization to provide auxiliary edge information. The most suitable $k$ is selected under the guidance of the one-dimensional structural entropy maximization strategy (\S~\ref{step1}). 
Upon the enhanced graph, we present a hierarchical abstraction mechanism to further suppress the edge noise and reveal the high-level hierarchical community structure (encoding tree) (\S~\ref{step2}). 
A novel sampling-based approach is designed to build new graph topology from the encoding tree, particularly by restoring the edge connectivity from the tree hierarchy (\S~\ref{step3}). 
The core idea is to weaken the association between high-level communities whilst establishing dense and extensive connections within low-level communities. 
To this end, we transform the node structural entropy into probability, rejecting the deterministic threshold. 
Through multi-iterative stochastic sampling, it is more likely to find favorable graph structures for GNNs. 
Afterward, the rebuilt graph will be fed into the downstream generic GNN encoders. To constantly improve both the node representation and the graph structure, the optimization pipeline is iterated for multiple epochs.
% -- the optimized node representation and graph structure will go into the next epoch as the new input. 
The training procedure of\ \framework{}\ is summarized in Appendix~\ref{appendix:overall algorithm}.

%介绍1维结构熵图强化
\subsection{Graph Structure Enhancement}\label{step1}

To fully incorporate vertex attributes and neighborhood information in the graph structure, we perform feature fusion and edge reweighting so that the topological structure, together with the informative vertex adjacent similarity, can be passed on to the encoding tree generator. 
To begin with, we calculate the pair-wise similarity matrix $S \in \mathbb{R}^{|V|\times |V|}$ among vertices in graph $G$. 
To better depict the linear correlation between two vertex attributes, we take the Pearson correlation coefficient (PCC) as the similarity measure in the experiments, i.e.,
\begin{equation}\label{eq:pcc}
S_{ij}=\mathrm{PCC}(x_i,x_j)=\frac{E((x_i-u_i)(x_j-u_j))}{\sigma_i\sigma_j},
\end{equation}
where $x_i$ and $x_j \in \mathbb{R}^{1 \times d}$ are the attribute vectors of vertices $i$ and $j$, respectively. $u_i$ and $\sigma_i$ denote the mean value and variance  of $x_i$, and $E(\cdot)$ is the dot product function.
Based on $S$, we can intrinsically construct the $k$-NN graph $G_{knn} = \{V,E_{knn}\}$ where each edge in $E_{knn}$ represents a vertex and its $k$ nearest neighbors (e.g., the edges in red in Fig ~\ref{fig:framework}). We fuse $G_{knn}$ and the original $G$ to $G_{f} = \{V,E_{f} = E \cup E_{knn}\}$.  

A key follow-up step is pinpointing the most suitable number $k$ of nearest neighbors. An excessive value of $k$ would make $G_{f}$ over-noisy and computationally inefficient, while a small $k$ would result in insufficient information and difficulties in hierarchy extraction. As outlined in \S~\ref{prelim:1dse}, a larger one-dimensional structural entropy indicates more information that $G_{f}$ can potentially hold. Hence, we aim to maximize the one-dimensional structural entropy $H^1(G_{f})$ to guide the selection of $k$ for larger encoding information. In practice, we gradually increase the integer parameter $k$, generate the corresponding $G^{(k)}_{f}$ and compute  $H^1(G^{(k)}_{f})$. Observably, when $k$ reaches a threshold $k_m$, $H^1(G^{(k)}_{f})$ comes into a plateau without noticeable increase. This motivates us to regard this critical point $k_m$ as the target parameter. The $k$-selector algorithm is depicted in Appendix ~\ref{appendix:1dse algorithm}. 
In addition, the edge $e_{ij}$ between $v_i$ and $v_j$ is reweighted as: 
\begin{equation}\label{eq:reweighted}
    \omega_{ij}=S_{ij}+M, \quad M = \frac{1}{2|V|} \cdot \frac{1}{|E|}\sum_{1<i,j<n}{S_{ij}}, 
\end{equation}
where $M$ is a modification factor that amplifies the trivial edge weights and thus makes the $k$-selector more sensitive to noises.

\subsection{Hierarchical Encoding Tree Generation}\label{step2}

Our methodology of abstracting the fused graph into a hierarchy is inspired by the structural entropy theory~\cite{li2016structural,li2018decoding}. We intend to minimize the graph uncertainty (i.e., edge noises) and maximize the knowledge embedded (e.g., optimal partition) in the high-dimensional hierarchy. Correspondingly, the objective of structural entropy minimization is to find out an encoding tree $\mathcal{T}$ that minimizes $H^\mathcal{T}(G_f)$ defined in Eq.~\ref{eq:HT}.
Due to the difficulty in graph semantic complexity quantification, we restrict the optimization objective to the $K$-level tree with a hyperparameter $K$. The optimal $K$-dimensional encoding tree is represented as:
\begin{equation}\label{eq:T}
\mathcal{T^*} = \mathop{\arg\min}\limits_{\forall\mathcal{T}:height(\mathcal{T})\le K}(H^\mathcal{T}(G)).
\end{equation}

To address this optimization problem, we design a greedy-based heuristic algorithm to approximate $H^K(G)$. To assist the greedy heuristic, we define two basic operators:

\begin{define}
\label{def:CBop}
\textbf{Combining operator:}~
Given an encoding tree $\mathcal{T}$ for $G=(V,E)$, let $\alpha$ and $\beta$ be two nodes in $\mathcal{T}$ sharing the same parent $\gamma$. 
The combining operator $\mathrm{CB}_{\mathcal{T}}(\alpha,\beta)$ updates the encoding tree as: $\gamma \gets \delta^-; \delta \gets \alpha^-; \delta \gets \beta^-.$ 
A new node $\delta$ is inserted between $\gamma$ and its children $\alpha, \beta$.
\end{define}
\begin{define}
\label{def:LFop}
\textbf{Lifting operator:}~
Given an encoding tree $\mathcal{T}$ for $G=(V,E)$, let $\alpha$, $\beta$ and $\gamma$ be the nodes in $\mathcal{T}$, satisfying $\beta^-=\gamma$ and $\alpha^-=\beta$.
The lifting operator $\mathrm{LF}_{\mathcal{T}}(\alpha,\beta)$ updates the encoding tree as:  $\gamma \gets \alpha^-; \mathrm{IF:}T_{\beta}=\varnothing, \mathrm{THEN:}\mathrm{drop}(\beta).$
The subtree rooted at $\alpha$ is lifted by placing itself as $\gamma$'s child. If no more children exist after lifting, $\beta$ will be deleted from $\mathcal{T}$.
\end{define}

In light of the high-dimensional structural entropy minimization principle~\cite{li2018decoding}, we first build a full-height binary encoding tree by greedily performing the combining operations. 
Two children of the root are combined to form a new partition iteratively until the structural entropy is no longer reduced. 
To satisfy the height restriction, we further squeeze the encoding tree by lifting subtrees to higher levels. 
To do so, we select and conduct lifting operations between a non-root node and its parent node that can reduce the structural entropy to the maximum.
This will be repeated until the encoding tree height is less than $K$ and the structural entropy can no longer be decreased.
Eventually, we obtain an encoding tree with a specific height $K$ with minimal structural entropy. The pseudo-code is detailed in Appendix~\ref{appendix:kdse algorithm}.

\subsection{Sample-based Graph Reconstruction}
\label{step3}

The subsequent step is to restore the topological structure from the hierarchy whilst guaranteeing the established hierarchical semantics in optimal encoding tree $\mathcal{T}^*$. 
The key to graph reconstruction is determining which edges to augment or weaken. 
Intuitively, the nodes in real-life graphs in different communities tend to have different labels. The work~\cite{zhu2020cagnn} has demonstrated the effectiveness of strengthening intra-cluster edges and reducing inter-cluster edges in a cluster-awareness approach to refine the graph topology. However, for hierarchical communities, simply removing cross-community edges will undermine the integrity of the higher-level community. Adding edges within communities could also incur additional edge noises to lower-level partitioning. 

We optimize the graph structure with community preservation by investigating the structural entropy of \textit{deduction} between two interrelated nodes as the criterion of edge reconstruction:

\begin{define}
\label{def:deduct-se}
\textbf{Structural entropy of deduction:}~
Let $\mathcal{T}$ be an encoding tree of $G$. We define the structural entropy of the deduction from non-leaf node $\lambda$ to its descendant $\alpha$ as:
\begin{equation}\label{eq:deduct-se}
% H^{\mathcal{T}}(G;(\lambda,\alpha]) = \sum_{\beta,\lambda \subset \beta \subseteq \alpha}{H^{\mathcal{T}}(G;\beta)}.
H^{\mathcal{T}}(G;(\lambda,\alpha]) = \sum_{\beta,T_\alpha \subseteq T_\beta \subset T_\lambda}{H^{\mathcal{T}}(G;\beta)}.
% \vspace{-0.4em}
\end{equation}
 This node structure entropy definition exploits the hierarchical structure of the encoding tree and offers a generic measure of top-down deduction to determine a community or vertex in the graph.
\end{define}

From the viewpoint of message passing, vertices with higher structural entropy of deduction produce more diversity and uncertainty and thus require more supervisory information.
Therefore, such vertices need expanded connection fields during the graph reconstruction to aggregate more information via extensive edges.
To achieve this, we propose an analog sampling-based graph reconstruction method. 
The idea is to explore the node pairs at the leaf node level (the lowest semantic level) and stochastically generate an edge for a given pair of nodes with a certain probability associated with the deduction structural entropy. 

Specifically, for a given $\mathcal{T}$, assume the node $\delta$ has a set of child nodes $\{ \delta^{\left \langle 1 \right \rangle},\delta^{\left \langle 2 \right \rangle},\dots,\delta^{\left \langle n \right \rangle}\} $. The probability of the child $\delta^{\left \langle i \right \rangle}$ is defined as: 
$P(\delta^{\left \langle i \right \rangle}) = \sigma_\delta(H^{\mathcal{T}}(G_{f};(\lambda,\delta^{\left \langle i \right \rangle}]))$,
% \begin{equation}\label{eq:prob}
% P(\delta^{\left \langle i \right \rangle}) = \sigma_\delta(H^{\mathcal{T}}(G_{f};(\lambda,\delta^{\left \langle i \right \rangle}])),
% \end{equation}
where $\lambda$ is the root of $\mathcal{T}$ and $\sigma_\delta(\cdot)$ represents a distribution function. Take $\mathrm{softmax}$ function as an example, the probability of $\delta^{\left \langle i \right \rangle}$ can be calculated as: 
\begin{equation}\label{eq:prob-softmax}
% \vspace{-0.4em}
P(\delta^{\left \langle i \right \rangle}) = \frac{\mathrm{exp}(H^{\mathcal{T}}(G_{f};(\lambda,\delta^{\left \langle i \right \rangle}]))}
{ {\textstyle \sum_{j=1}^{n}}{\mathrm{exp}(H^{\mathcal{T}}(G_{f};(\lambda,\delta^{\left \langle j \right \rangle}]))} }.
\end{equation}

The probability of a non-root node can be acquired recursively. To generate new edges, we sample leaf node pairs by a top-down approach with a single sampling flow as follows: 

\noindent \textbf{(1)} For the encoding tree (or subtree) with root node $\delta$, two different child nodes $\delta^{\left \langle i \right \rangle}$ and $\delta^{\left \langle j \right \rangle}$ are selected by sampling according to $P(\delta^{\left \langle i \right \rangle})$ and $P(\delta^{\left \langle j \right \rangle})$. Let $\delta_1 \gets \delta^{\left \langle i \right \rangle}$ and $\delta_2 \gets \delta^{\left \langle j \right \rangle}$
\textbf{(2)} If $\delta_1$ is a non-leaf node, we perform sampling once on the subtree rooted at $\delta_1$ to get $\delta_1^{\left \langle i \right \rangle}$, then update $\delta_1 \gets \delta_1^{\left \langle i \right \rangle}$. The same is operated on $\delta_2$.
\textbf{(3)} After recursively performing step (2), we sample two leaf nodes $\delta_1$ and $\delta_2$, while adding the edge connecting vertex $v_1 = T_{\delta_1}$ and $v2 = T_{\delta_2}$ into the edge set $E'$ of graph $G'$. 
To establish extensive connections at all levels, multiple samplings are performed on all encoding subtrees. 
For each subtree rooted at $\delta$, we conduct independent samplings for $\theta \times n$ times, where $n$ is the number of $\delta$'s children, and $\theta$ is a hyperparameter that positively correlated with the density of reconstructed graph. 
For simplicity, we adopt a uniform $\theta$ for all subtrees. 
% If it is necessary to precisely control the sparsity within communities and the connectivity between communities, 
Separately setting and tuning $\theta$ of each semantic level for precise control is also feasible.

\subsection{Time Complexity of \ \framework{}}\label{timecomplexity}
The overall time complexity is $O(n^2+n+n\log ^2n)$, in which $n$ is the number of nodes. 
Separately, in \S ~\ref{step1}, the time complexity of calculating similarity matrix is $O(n^2)$ and of $k$-selector is $O(n)$. 
According to ~\cite{li2016structural}, the optimization of a high-dimensional encoding tree in \S ~\ref{step2} costs the time complexity of $O(n\log ^2n)$.
As for the sampling process in \S ~\ref{step3}, the time complexity can be proved as $O(2n)$.
We report the time cost of \ \framework{} in Appendix~\ref{appendix:baseline}.
%% 3-purposed method
\section{Experimental Setup}\label{sec:experi}

\begin{table*}[htb!]
    \renewcommand{\arraystretch}{1.05}
    \setlength{\abovecaptionskip}{0.25cm}
    \setlength{\belowcaptionskip}{-0.25cm}
    % \caption{Classification Accuracy (\%) comparison, with improvement range of ~\framework~ against the baselines.}
    \caption{Classification Accuracy (\%) comparison, with improvement range of ~\framework~ against the baselines. \textmd{The best results are bolded and the second-best are underlined. \posimp{Green} denotes the outperformance percentage, while \negimp{yellow} denotes underperformance.}}
    \label{tab:performance comparison}
\centering
    % \scalebox{1.0}{
    \setlength{\tabcolsep}{1.2mm}{
        % \begin{tabular}{c|c|c|c|c|c|c|c|c|c|c}
        \begin{tabular}{l|ccccccccc}
        \toprule
        %\multirow{2}*{\textbf{Blocks}} & \multicolumn{2}{c|}{\textbf{Traditional}} & \multicolumn{4}{c|}{\textbf{Evolutionary}} & \multicolumn{5}{c}{\textbf{Dedicated}} \\
        %\cline{2-13}
       Method & {Cora} & {Citeseer} & {Pubmed} & {PT} & {TW} & {Actor} & {Cornell} & {Texas} & {Wisconsin} \\
        % \cline{1-1}
        %Method & & & & & & & & &\\
        \toprule     % cora    cite    pub     squ    cham    act      cor      tex     wis
        GCN          & 87.26$_{\pm0.63}$ & 76.22$_{\pm0.71}$ & 87.46$_{\pm0.12}$ & 67.62$_{\pm0.21}$ & 62.46$_{\pm1.94}$
                     & 27.65$_{\pm0.55}$ & 49.19$_{\pm1.80}$ & 57.30$_{\pm2.86}$ & 48.57$_{\pm4.08}$\\
        GAT          & 87.52$_{\pm0.69}$ & 76.04$_{\pm0.78}$ & 86.61$_{\pm0.15}$ & 68.76$_{\pm1.01}$ & 61.68$_{\pm1.20}$
                     & 27.77$_{\pm0.59}$ & 57.09$_{\pm6.32}$ & 58.10$_{\pm4.14}$ & 51.34$_{\pm4.78}$\\
        GCNII        & 87.57$_{\pm0.87}$ & 75.47$_{\pm1.01}$ & \underline{88.64$_{\pm0.23}$} & 68.93$_{\pm0.93}$ & 65.17$_{\pm0.47}$ 
                     & 30.66$_{\pm0.66}$ & 58.76$_{\pm7.11}$ & 55.36$_{\pm6.45}$ & 51.96$_{\pm4.36}$\\
        Grand        & \textbf{87.93$_{\pm0.71}$} & 77.59$_{\pm0.85}$ & 86.14$_{\pm0.98}$ & 69.80$_{\pm0.75}$ & \underline{66.79$_{\pm0.22}$}  
                     & 29.80$_{\pm0.60}$ & 57.21$_{\pm2.48}$ & 56.56$_{\pm1.53}$ & 52.94$_{\pm3.36}$ \\
        %Mixhop       & & & & & & & & & \\
        \midrule    % cora    cite     pub     squ    cham    act     cor     tex      wis
        % Geom-GCN-I   & 85.19 & 77.99 & 90.05 & - & - & 29.09 & 56.76 & 57.58 & 58.24 \\
        Mixhop     & 85.71$_{\pm0.85}$ & 75.94$_{\pm1.00}$ & 87.31$_{\pm0.44}$ & 69.48$_{\pm0.30}$ & 66.34$_{\pm0.22}$ & 33.72$_{\pm0.76}$ & 64.47$_{\pm4.78}$ & 63.16$_{\pm6.28}$ & 72.12$_{\pm3.34}$ \\
        Dropedge   & 86.32$_{\pm1.09}$  & 76.12$_{\pm1.32}$  & 87.58$_{\pm0.34}$  & 68.49$_{\pm0.91}$  & 65.24$_{\pm1.45}$  & 30.10$_{\pm0.71}$  & 58.94$_{\pm5.95}$ & 59.20$_{\pm5.43}$ & 60.45$_{\pm4.48}$ \\
        Geom-GCN-P   & 84.93 & 75.14 & 88.09 & - & - & 31.63 & 60.81 & 67.57 & 64.12 \\
        Geom-GCN-S   & 85.27 & 74.71 & 84.75 & - & - & 30.30 & 55.68 & 59.73 & 56.67 \\
        GDC          & 87.17$_{\pm0.36}$ & 76.13$_{\pm0.53}$ & 88.08$_{\pm0.16}$ & 66.14$_{\pm0.54}$ & 64.14$_{\pm1.40}$ 
                     & 28.74$_{\pm0.76}$ & 59.46$_{\pm4.35}$ & 56.42$_{\pm3.99}$ & 48.30$_{\pm4.29}$ \\ 
        % Pro-GNN      & 83.98$_{\pm0.86}$ & 72.41$_{\pm0.99}$ & 88.56$_{\pm0.37}$ & 35.89$_{\pm1.37}$ & 57.04$_{\pm2.20}$ 
                    %  & 27.96$_{\pm1.74}$ & 54.35$_{\pm6.79}$ & 56.76$_{\pm6.02}$ & 52.94$_{\pm5.88}$ \\
        GEN          & \underline{87.84$_{\pm0.69}$} & \textbf{78.77$_{\pm0.88}$} & 86.13$_{\pm0.41}$ & \underline{71.62$_{\pm0.78}$} & 65.16$_{\pm0.77}$
                     & \textbf{36.69$_{\pm1.02}$} & 65.57$_{\pm6.74}$ & 73.38$_{\pm6.65}$ & 54.90$_{\pm4.73}$ \\
        H$_2$GCN-2   & 87.81$_{\pm1.35}$ & 76.88$_{\pm1.77}$ & \textbf{89.59$_{\pm0.33}$} & 68.15$_{\pm0.30}$ & 63.33$_{\pm0.77}$
                     & 35.62$_{\pm1.30}$ & \textbf{82.16$_{\pm6.00}$} & \underline{82.16$_{\pm5.28}$} & \underline{85.88$_{\pm4.22}$} \\
        \midrule     
        % \framework   & 89.32$_{\pm1.24}$ & 78.85$_{\pm2.35}$ & 88.86$_{\pm0.75}$ & 36.07$_{\pm3.42}$ & 56.98$_{\pm2.81}$ 
        %              & 36.20$_{\pm2.07}$ & 76.85$_{\pm5.89}$ & 84.49$_{\pm4.80}$ & 86.27$_{\pm4.32}$ \\
        \framework   & \textbf{87.93$_{\pm1.24}$} & \underline{77.63$_{\pm1.65}$} & 88.16$_{\pm0.76}$ & \textbf{71.91$_{\pm0.66}$} & \textbf{66.99$_{\pm0.93}$} 
                     & \underline{36.34$_{\pm2.07}$} & \underline{75.21$_{\pm5.54}$} & \textbf{82.49$_{\pm4.80}$} & \textbf{86.27$_{\pm4.32}$} \\
        \midrule      
        Improvement  & \posimp{0.00}$\sim$\posimp{3.00}  & \negimp{-1.14}$\sim$\posimp{2.92}  & \negimp{-1.43}$\sim$\posimp{3.41}
                     & \posimp{0.29}$\sim$\posimp{5.77} & \posimp{0.20}$\sim$\posimp{5.31} &  \negimp{-0.35}$\sim$\posimp{8.69}
                     & \negimp{-6.95}$\sim$\posimp{26.02} & \posimp{0.33}$\sim$\posimp{27.13} & \posimp{0.39}$\sim$\posimp{37.97}\\
        
        \bottomrule
        % DGI          & 82.72 & XXXXX & XXXXX & 33.69 & 52.27 & 25.96 & 54.09 & 58.11 & 48.04\\
        % Dropedge(GCN)& XXX & XXX & XXX & XXX & 45.36 & 68.64 & XXX & 57.89 & XXX & XXX\\
        \end{tabular}
    }
\end{table*}

\noindent \textbf{Software and Hardware.}  
All experiments are conducted on a Linux server with GPU (NVIDIA RTX A6000) and CPU (Intel i9-10980XE), using PyTorch 1.12 and Python 3.9.

\noindent \textbf{Datasets}. 
We experiment on nine open graph benchmark datasets, including three citation networks (i.e., Cora, Citeseer, and Pubmed), two social networks (i.e., PT and TW), three website networks from WebKB (i.e., Cornell, Texas, and Wisconsin), and a co-occurrence network. 
Their statistics are summarized in Appendix ~\ref{appendix:dataset}.

\noindent \textbf{Baseline and backbone models}. 
We compare ~\framework{} with baselines including general GNNs (i.e., GCN, GAT, GCNII, Grand) and graph learning/high-order neighborhood awareness methods (i.e. MixHop, Dropedge, Geom-GCN, GDC, GEN, H$_2$GCN). 
Four classic GNNs (GCN, GAT, GraphSAGE, APPNP) are chosen as the backbone encoder that ~\framework{} works upon. Details are in Appendix ~\ref{appendix:baseline}.

\noindent \textbf{Parameter settings}. 
For ~\framework~ with various backbones, we uniformly adopt two-layer GNN encoders. 
To avoid over-fitting, We adopt ReLU (ELU for GAT) as the activation function and apply a dropout layer with a dropout rate of 0.5. The training iteration is set to 10.
The embedding dimension $d$ is chosen from $\{8,16,32,48,64\}$, while the height of the encoding tree $K$ is searched among $\{2,3,4\}$, and the hyperparameter $\theta$ in \ref{step3} is tuned among $\{0.5, 1, 3, 5, 10,30\}$. 
We adopt the scheme of data split in Geom-GCN~\cite{pei2019geom} and H$_2$GCN~\cite{zhu2020beyond} for all experiments -- nodes are randomly divided into the train, validation, and test sets, which take up 48\%, 32\%, 20\%, respectively.
In each iteration, the GNN encoder optimization is carried out for 200 epochs, using the Adam optimizer, with an initial learning rate of $0.01$ and a weight decay of $5e-4$. The model with the highest accuracy on validation sets is used for further testing and reporting.

\begin{figure*}[htb]
  \centering
  \includegraphics[width=0.99\textwidth]{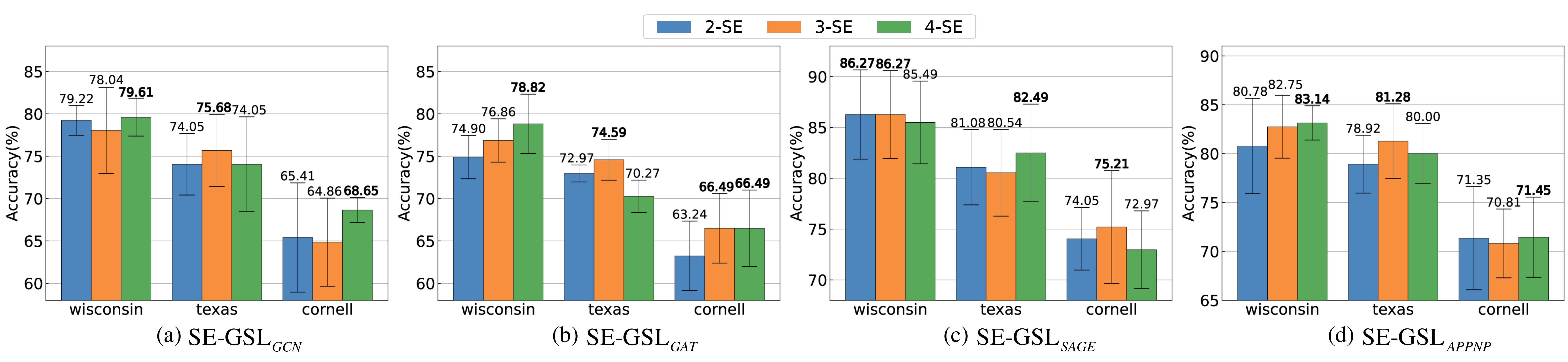}
  \caption{Results of ~\framework{} with different encoding tree heights.}
  \label{fig:K-bar}
  \Description{Results with different encoding tree heights.}
\end{figure*}

\section{Results and Analysis}\label{sec:evalu}
In this section, we demonstrate the efficacy of \ \framework{} on semi-supervised node classification (\S ~\ref{sec:exp:overall}, followed by micro-benchmarks that investigate the detailed effect of the submodules on the overall performance and validate the robustness of ~\framework{} when tackling random perturbations (\S ~\ref{sec:exp:micro}). For better interpretation, we visualize the change of structural entropy and graph topology (\S ~\ref{sec:exp:int}).

\subsection{Node Classification}
\label{sec:exp:overall}

\subsubsection{Comparison with baselines}
We compare the node classification performance of ~\framework~ with ten baseline methods on nine benchmark datasets.
Table ~\ref{tab:performance comparison} shows the average accuracy and the standard deviation. 
Note that the results of H$_2$GCN (except PT and TW) and Geom-GCN are from the reported value in original papers ( - for not reported), while the rest are obtained based on the execution of the source code provided by their authors under our experimental settings. Our observations are three-fold: 
% \textbf{(1)} While all GNN methods can achieve satisfactory results on citation networks, graph structural learning frameworks perform significantly better than conventional GNN methods on WebKB, Wiki, and actor co-occurrence networks due to the high heterophily of these networks.
% The reason for this phenomenon is these conventional GNN models agggregate to much classification-irrelevant information from disassortative neighborhoods. 
% In contrast, graph structural learning can optimize the neighborhood topology to achieve better results.

% \textbf{(1)} While all GNN methods can achieve satisfactory results on citation networks, the specialized graph learning frameworks perform significantly better on WebKB, Wiki and actor co-occurence networks due to the heterophily challenge. 

\noindent \textbf{(1)} \framework~ achieves optimal results on 5 datasets, runner-up results on 8 datasets, and advanced results on all datasets. The accuracy can be improved up to 3.41\% on Pubmed, 3.00\% on Cora, and 2.92\% on Citeseer compared to the baselines. This indicates that our design can effectively capture the inherent and deep structure of the graph and hence the classification improvement.

\noindent \textbf{(2)} \framework~ shows significant improvement on the datasets with heteropily graphs, e.g., up to 37.97\% and 27.13\% improvement against Wisconsin and Texas datasets, respectively. This demonstrates the importance of the graph structure enhancement that can contribute to a more informative and robust node representation.

\noindent \textbf{(3)} While all GNN methods can achieve satisfactory results on citation networks, the graph learning/high-order neighborhood awareness frameworks substantially outperform others on the WebKB datasets and the actor co-occurrence networks, which is highly disassortative. This is because these methods optimize local neighborhoods for better information aggregation. Our method is one of the top performers among them due to the explicit exploitation of the global structure information in the graph hierarchical semantics.

\subsubsection{Comparison base on different backbones}
Table~\ref{tab:backbone comparison} shows the mean classification accuracy of ~\framework{} with different backbone encoders.
Observably, ~\framework{} upon GCN and GAT overwhelmingly outperforms its backbone model, with an accuracy improvement of up to 31.04\% and 27.48\%, respectively. 
This indicates the iterative mechanism in the ~\framework{} pipeline can alternately optimize the node representation and graph structure.
We also notice that despite the lower improvement, ~\framework{} variants based on GraphSAGE and APPNP perform relatively better compared to those on GCN and GAT.
This is most likely due to the backbone model itself being more adapted to handle disassortative settings on graphs.
% We also notice that ~\framework~ based on GraphSAGE has the lowest improvement. This is most likely due to the weak adaptability of the backbone model itself to disassortative settings.

\begin{table}[t]
    \renewcommand{\arraystretch}{1.05}
    \setlength{\abovecaptionskip}{0.15cm}
    \setlength{\belowcaptionskip}{-0.25cm}
    \caption{Classification accuracy(\%) of ~\framework{} and corresponding backbones. Wisc. is short for Wisconsin.}%mean relative 
    \label{tab:backbone comparison}
    \centering
    % \scalebox{0.9}{
    % \setlength{\tabcolsep}{1mm}{
        \begin{tabular}{l|ccccc}
        \hline
        Method & Actor & TW & Texas & Wisc. & Improvement\\
        \hline
        \framework$_{GCN}$   & 35.03 & 66.88 & 75.68 & 79.61 & $\uparrow$ 5.20$\sim$31.04\\
        % Chebnet         &
        % \framework(Chebnet)
        % \midrule
        % SGC             &
        % \framework(SGC)
        % \framework
        % (SAGE)            & 36.34 & 66.92 & \textbf{81.62} & \textbf{86.27} & $\uparrow$ 0.25$\sim$3.79\\
          \framework$_{SAGE}$& 36.20 & 66.92 & \textbf{82.49} & \textbf{86.27} & $\uparrow$ 0.25$\sim$6.79\\
        \framework$_{GAT}$   & 32.46  & 63.57 & 74.59 & 78.82 & $\uparrow$ 4.69$\sim$27.48\\
        % \framework(APPNP) & \textbf{36.62} & 71.45 & 81.28 & 83.14 & 13.34\%\\
        \framework$_{APPNP}$ & \textbf{36.34} & \textbf{66.99} & 81.28 & 83.14 & $\uparrow$ 2.01$\sim$12.16\\
        \hline
        \end{tabular}
    % }
    % }
\end{table}

\begin{table}[t]
    \setlength{\abovecaptionskip}{0.15cm}
    \setlength{\belowcaptionskip}{-0.25cm}
    \caption{The $k$ selection for each iteration in structural optimization. Bolds represent the $k$ selection when the accuracy reaches maximum.}\label{tab:k-NN comparison}
    \centering
    % \scalebox{0.9}{
    \setlength{\tabcolsep}{2mm}{
        \begin{tabular}{l|ccccccccc}
        \hline
        Iteration & 1 & 2 & 3 & 4 & 5 & 6 & 7 & 8 & 9\\
        \hline
        {Cora}   & 22 & \bf{22} & 19 & 22 & 21 & 22 & 20 & 21 & 20\\
        {Actor}  & 23   & 15 & 15 & 15 & 14 & 15 & 14 & \bf{14} & 15\\
        {TW} & 50 & 16 & 16 & \bf{17} & 15 & 17 & 27 & 16 & 16 \\
        {Wisconsin} & 21 & 16 & \bf{11} & 16 & 14 & 13 & 16 & 13 & 11 \\
        {Texas}  & 21 & 13 & 13 & \bf{13} & 13 & 10 & 14 & 10 & 14 \\
        \hline
        \end{tabular}
    }
\end{table}

\subsection{Micro-benchmarking}
\label{sec:exp:micro}

\subsubsection{Effectiveness of~$k$-selector}
This subsection evaluate how the one-dimensional structural entropy guides the $k$-selector in \S ~\ref{step1}.
Table ~\ref{tab:k-NN comparison} showcases the selected parameter $k$ in each iteration with \framework$_{GCN}$. 
Noticeably, as the iterative optimization proceeds, the optimal parameter $k$ converges to a certain range, indicating the gradual stabilization of the graph structure and node representation. The disparity of parameter $k$ among different datasets also demonstrates the necessity of customizing $k$ in different cases rather than using $k$ as a static hyperparameter.

\subsubsection{Impact of the encoding tree's height $K$}
We evaluate all four variants of ~\framework~ on the website network datasets, and the encoding tree height $K$ involved in \S ~\ref{step2} varies from 2 to 4.
As shown in Fig. ~\ref{fig:K-bar}, there is a huge variation in the optimal tree heights among different datasets. For example, in the variants based on GAT, GCN, and APPNP, the best results can be targeted at $K=3$ in Texas and at $K=4$ in Cornell and Wisconsin. By contrast, in ~\framework$_{SAGE}$,  $K=2$ can enable the best accuracy of 86.27\%. This weak correlation between the best $K$ and the model performance is worth investigating further, which will be left as future work.

\begin{figure}[tb]
  \centering
  \includegraphics[width=0.48\textwidth]{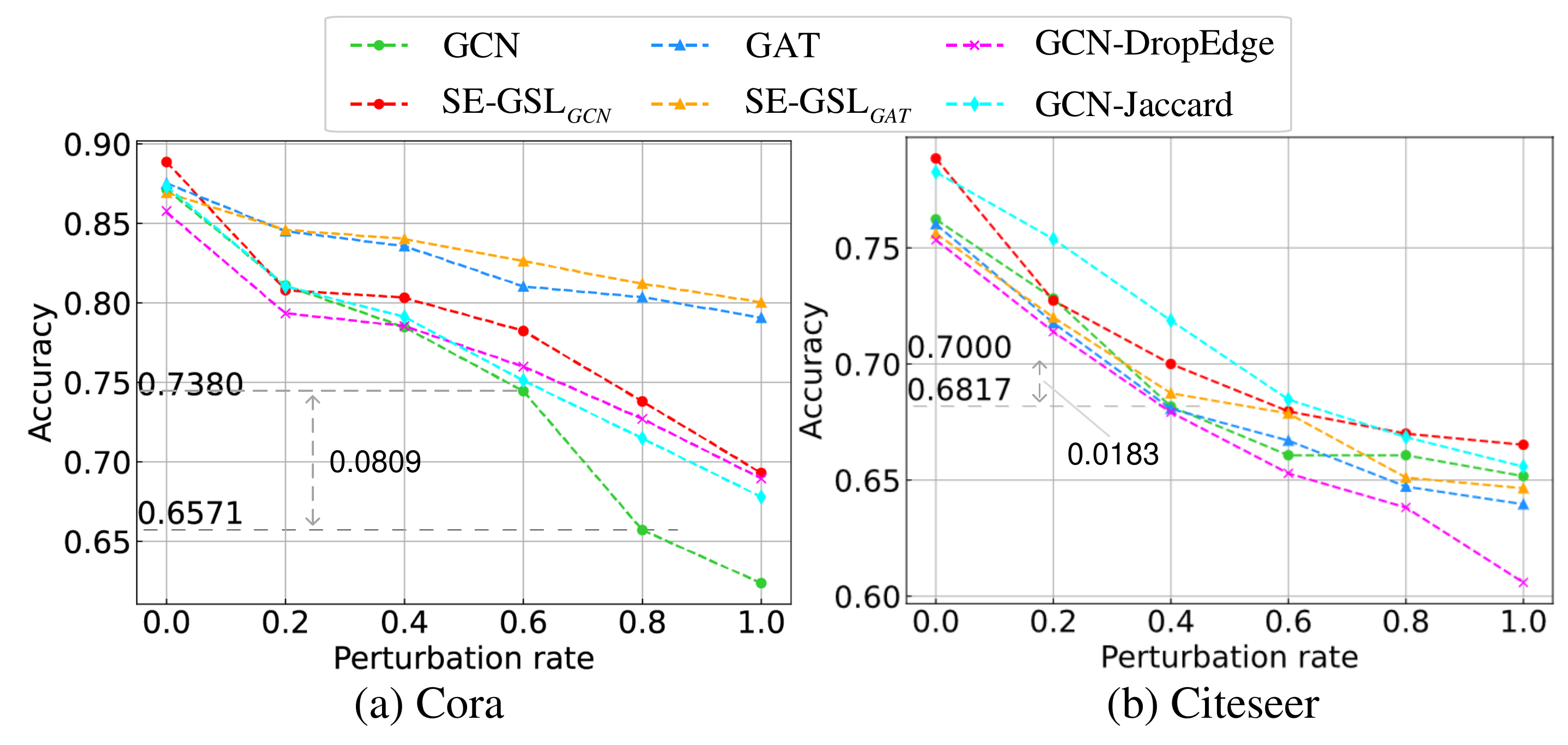}
  \caption{Robustness of ~\framework~ against random noises.}
  \label{fig:pertubation}
  \Description{Results of perturbation experiment.}
\end{figure}

\begin{figure*}[tb]
  \centering
  \includegraphics[width=0.99\textwidth]{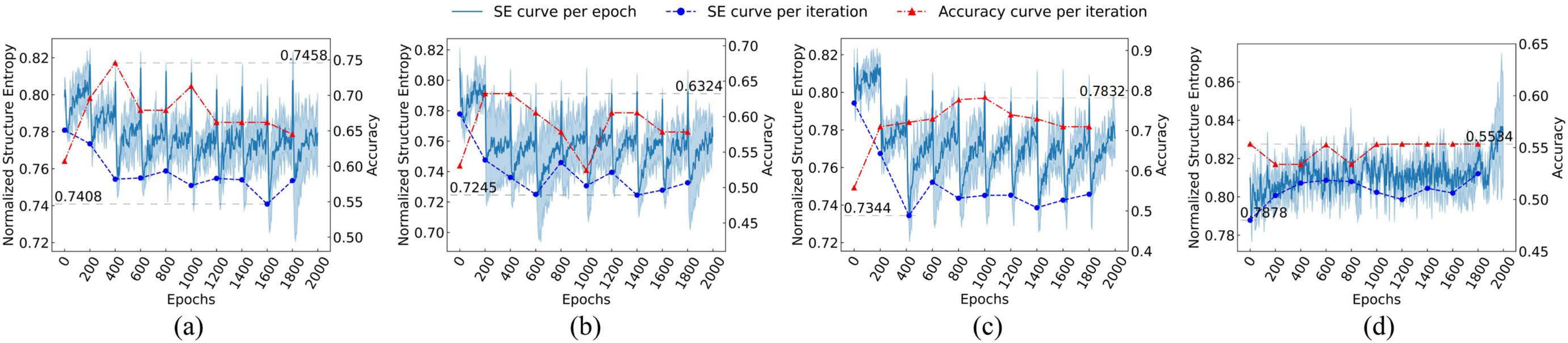}
  \caption{The normalized structural entropy changes during the training of ~\framework$_{GAT}$ with 2-dimensional structural entropy on (a) Texas, (b) Cornell, and (c) Wisconsin. The structure is iterated every 200 epochs. By comparison, (d) shows the entropy changes on Wisconsin without the graph reconstruction strategy.}
  \label{fig:sedecrease}
  \Description{Visualization of structural entropy and acc. variation.}
\end{figure*}

\begin{figure}[tb]
  \centering
  \includegraphics[width=0.49\textwidth]{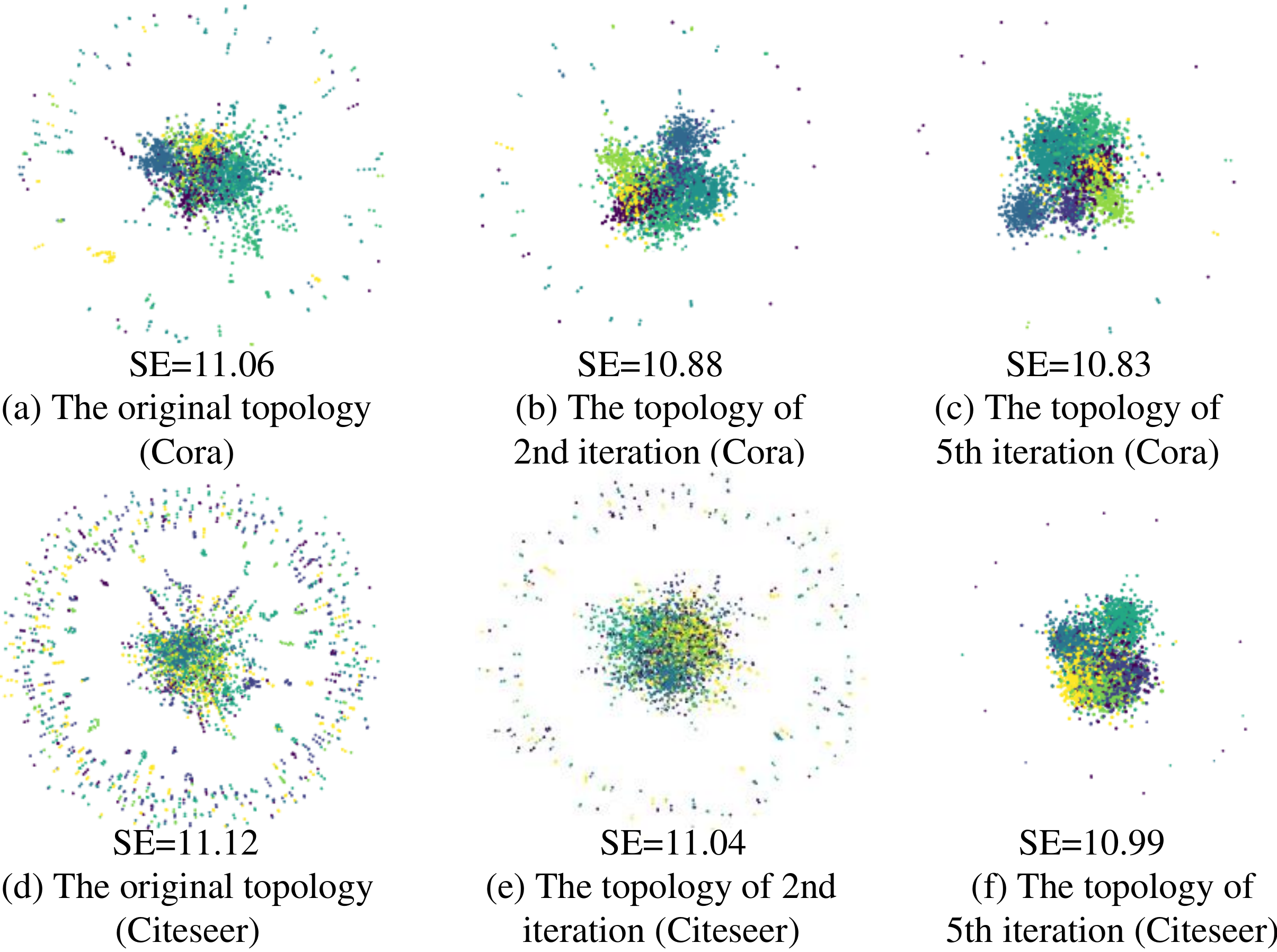}
  \caption {The visualized evolution of the graph structure on Cora (a,b,c) and Citeseer (d,e,f). The corresponding Structural Entropy (SE) is also shown.}
  \label{fig:topovisual}  
  \Description{Visualization of topology evolution.}
\end{figure}

\subsubsection{Sensitivity to perturbations}
We introduce random edge noises into Cora and Citeseer, with perturbation rates of 0.2, 0.4, 0.6, 0.8, and 1. As shown in Fig.~\ref{fig:pertubation}(a), ~\framework{} outperforms baselines in both GCN and GAT cases under most noise settings. For instance, ~\framework$_{GCN}$ achieves up to 8.09\% improvement against the native GCN when the perturbation rate is 0.8; by contrast, improvements by GCN-Jaccard and GCN-DropEdge are merely 6.99\% and 5.77\%, respectively. A similar phenomenon is observed for most cases in the Citeseer dataset (Fig.~\ref{fig:pertubation}(b)), despite an exception when compared against GCN-Jaccard. Nevertheless, our approach is still competitive and even better than GCN-Jaccard at a high perturbation rate. 

\subsection{Interpretation of Structure Evolution}
\label{sec:exp:int}

\subsubsection{Structural entropy variations analysis}
We evaluate how the structural entropy changes during the training of ~\framework$_{GAT}$ with 2-dimensional structural entropy on WebKB datasets. For comparison, we visualize the entropy changes on Wisconsin without the structure learning. In the experiment setting, both the graph structure and the encoding tree are updated once at each iteration (i.e., 200 GNN epochs), and within one iteration, the structural entropy is only affected by edge weights determined by the similarity matrix. For comparison, we normalize the structural entropy by $ \textstyle{\frac{H^{\mathcal{T}}(G)}{H^1(G)}}$.

As shown in Fig.~\ref{fig:sedecrease}(a)-(c), as the accuracy goes up, the normalized structural entropy constantly decreases during the iterative graph reconstruction, reaching the minimums of 0.7408 in Texas, 0.7245 in Cornell, and 0.7344 in Wisconsin. This means the increasing determinism of the overall graph structure and the reduced amount of information required to determine a vertex. 
Interestingly, if our graph reconstruction mechanism is disabled (as shown in Fig.~\ref{fig:sedecrease}(d)), the normalized structural entropy keeps rising from 0.7878, compared with Fig.~\ref{fig:sedecrease}(c). Accordingly, the final accuracy will even converge to 55.34\%, a much lower level. 

Such a comparison also provides a feasible explanation for the rising trend of the normalized structural entropy within every single iteration. 
This stems from the smoothing effect during the GNN training. 
As the node representation tends to be homogenized, the graph structure will be gradually smoothed, leading to a decrease in the one-dimensional structural entropy thus the normalized structural entropy increases.
% We speculate that the continuous rise in structural entropy may be a feasible explanation for the over-smoothing of graph neural networks, which requires further study.
% In summary, this experiment well explains the robustness of our framework.

\subsubsection{Visualization}
Fig.~\ref{fig:topovisual} visualizes the topology of the original Cora and Citeseer graph and of the 2nd and 5th iterations.
The vertex color indicates the class it belongs to, and the layout denotes connecting relations. Edges are hidden for clarity. As the iteration continues, much clearer clustering manifests -- few outliers and more concentrated clusters.  
Vertices with the same label are more tightly connected due to the iterative graph reconstruction scheme. This improvement hugely facilitates the interpretability of the GSL and the node representation models. 

pan2021information\section{Related Work}\label{sec:relate}

\noindent {\textbf{Graph structure learning and neighborhood optimization.}}
The performance of GNNs is heavily dependent on task-relevant links and neighborhoods.
Graph structure learning explicitly learns and adjusts the graph topology, and our \framework{} is one of them.
GDC~\cite{gasteiger_diffusion_2019} reconnects graphs through graph diffusion to aggregate from multi-hop neighborhoods.
Dropedge~\cite{rong2019dropedge}, Neuralsparse ~\cite{zheng2020robust} contribute to graph denoising via edge-dropping strategy while failing to renew overall structures.
LDS-GNN~\cite{franceschi2019learning} models edges by sampling graphs from the Bernoulli distribution. Meanwhile, we consider linking the structural entropy, which is more expressive of graph topology, to the sampling probability.
GEN~\cite{wang2021graph}, IDGL~\cite{chen2020iterative} explore the structure from the node attribute space by the $k$-NN method. Differently, instead of directly using attribute similarity, we regenerate edges from the hierarchical abstraction of graphs to avoid inappropriate metrics.
Besides adjusting the graph structure, methods to optimize aggregation are proposed with results on heterophily graphs.
MixHop~\cite{abu2019mixhop} learns the aggregation parameters for neighborhoods of different hops through a mixing network, while H$_2$GCN~\cite{zhu2020beyond} 
% iteratively concatenates the multi-hop neighborhood features for the final node embeddings.
identifies higher-order neighbor-embedding separation and intermediate representation combination, for adapting to heterophily graphs.
Geom-GCN~\cite{pei2019geom} aggregates messages over both the original graph and latent space by a designed geometric scheme.

\noindent {\textbf{Structural entropy with neural networks.}}
Structural information principles~\cite{li2016structural}, defining encoding trees and structural entropy, were first used in bioinformatic structure analysis~\cite{li2016three,li2018decoding}. 
Existing work mainly focuses on network analysis, node clustering and community detection\cite{li2017resistance,liu2019rem,pan2021information}.
As an advanced theory on graphs and hierarchical structure, structural information theory has great potential in combination with neural networks. 
SR-MARL~\cite{zeng2023effective} applies structural information principles to hierarchical role discovery in multi-agent reinforcement learning, thereby boosting agent network optimization.
SEP~\cite{wu2022structural} provides a graph pooling scheme based on optimal encoding trees to address local structure damage and suboptimal problem. It essentially uses structural entropy minimization for a multiple-layer coarsened graph.
MinGE~\cite{luo2021graph} and MEDE~\cite{yang2023wsdm} estimate the node embedding dimension of GNNs via structural entropy, which introduces both attribute entropy and structure entropy as objective.
Although these works exploit structural entropy to mine the latent settings of neural networks and GNNs, how to incorporate this theory in the optimization process is still understudied, and we are among the first attempts.
%% note 
%Although plenty of works have attributed a natural hierarchical structure to graphs, an approach to the most justified and explainable partition has been the angle of discussion. 
\section{Conclusion}
\label{sec:conclu}

To cope with edge perturbations in graphs with heterophily, this paper proposes a novel graph structure learning framework ~\framework{} that considers the structural entropy theory in graph structure optimization.  
We design a structure enhancement module guided by the one-dimensional structural entropy maximization strategy to extend the information embedded in the topology. To capture the hierarchical semantics of graphs, high-dimensional structural entropy minimization is performed for optimal encoding trees. We propose a node sampling technique on the encoding tree to restore the most appropriate edge connections at different community levels, taking into account the deduction structural entropy distribution. In the future, we plan to combine delicate loss functions with structural entropy so that the knowledge in encoding trees can be converted into gradient information, which will further allow for end-to-end structure optimization.

\begin{acks}
This paper was supported by the National Key R\&D Program of China through grant 2021YFB1714800, NSFC through grant 62002007, S\&T Program of Hebei through grant 20310101D, Natural Science Foundation of Beijing Municipality through grant 4222030, CCF-DiDi GAIA Collaborative Research Funds for Young Scholars, the Fundamental Research Funds for the Central Universities, Xiaomi Young Scholar Funds for Beihang University, and in part by NSF under grants III-1763325, III-1909323,  III-2106758, and SaTC-1930941. 
\end{acks}

\bibliographystyle{ACM-Reference-Format}
\bibliography{sample-base}

\clearpage
\appendix
% \newpage
% \newpage
\section{Appendix}\label{sec:append}
\subsection{Glossary of notations}
\label{appendix:notations}
In Table ~\ref{tab:notations}, we summarize the notations used in our work. 

\begin{table}[ht]
    \setlength{\abovecaptionskip}{0.25cm}
    \setlength{\belowcaptionskip}{-0.25cm}
    \caption{Glossary of Notations.}\label{tab:notations}
    % \vspace{-2.5mm}
    \centering
    \scalebox{1.0}{
        \begin{tabular}{l|l}
        \hline
        Notation & Description \\
        \hline
        $G;A;S$ & Graph; Adjacency matrix; Similarity matrix. \\
        % $A^{G}_{ij}$ & The weight of the edge between $v_i$ and $v_j$ in $G$.\\
        $v;e;x$ & Vertex; Edge; Vertex attribute. \\
        $V;E;X$ & Vertex set; Edge set; Attribute set. \\
        $|V|;|E|$ & The number of vertices and edges. \\
        $\mathcal{P};P_i$ & The partition of $V$; A community.\\
        $D;d(v_i)$ & The degree matrix; The degree of vertex $v_i$. \\
        $e_{ij}$ & The edge between $v_i$ and $v_j$. \\
        $w_{ij}$ & The weight of edge $e_{ij}$. \\
        $vol(G)$ & The volume of graph $G$, i.e., degree sum in $G$. \\
        $G^{(k)}_{knn}$ & The $k$-NN graph with parameter $k$.\\
        $G_{f}$  & Fusion graph.\\
        $G^{(k)}_{f}$ & The fusion graph with parameter $k$.\\
        \hline
        $\mathcal{T}$ & Encoding tree. \\
        $\mathcal{T}^*$ & The optimal encoding tree. \\
        % $\mathcal{T}^{(K)}$ & The encoding tree with height $K$. \\
        $\lambda$ & The root node of an encoding tree. \\
        $\alpha$ & A non-root node of an encoding tree. \\
        $\alpha^-$ & The parent node of $\alpha$. \\
        $\alpha^{\left \langle i \right \rangle}$ & the $i$-th child of $\alpha$.\\
        $T_\lambda$ & The label of $\lambda$, i.e., node set $V$. \\
        $T_\alpha$ & The label of $\alpha$, i.e., a subset of $V$.\\
        $\mathcal{V}_\alpha$ & Volume of graph $G$. \\
        $g_a$ & the sum weights of cut edge set $[T_\alpha,T_\alpha/T_\lambda]$. \\
        $N(\mathcal{T})$ & The number of non-root node in $\mathcal{T}$.\\
        \hline
        $H^\mathcal{T}(G)$ & Structural entropy of $G$ under $\mathcal{T}$.\\
        $H^K(G)$ & $K$-dimensional structural entropy.\\
        $H^1(G)$ & One-dimensional structural entropy.\\
        $H^\mathcal{T}(G;\alpha)$ & Structural entropy of node $\alpha$ in $\mathcal{T}$.\\
        $H^{\mathcal{T}}(G;(\lambda,\alpha])$ & Structural entropy of a deduction from $\lambda$ to $\alpha$.\\
        \hline
        \end{tabular}
    }
% \vspace{-6.5mm}
\end{table}

% \vspace{-0.8em}
\subsection{Dataset and Time Costs of \ \framework{}}
\label{appendix:dataset}
Our framework \framework~ is evaluated on nine graph datasets. the statistics of these datasets are shown in Table~\ref{tab:statistics}. The time costs of \ \framework{} on all datasets are shown in Table~\ref{tab:time comparasion}.
% \vspace{-0.2em}
\begin{table}[ht]
    \setlength{\abovecaptionskip}{0.25cm}
    \setlength{\belowcaptionskip}{-0.25cm}
    \caption{Statistics of benchmark datasets.}\label{tab:statistics}
    \centering
    % \vspace{-0.2em}
    \scalebox{1.0}{
        \begin{tabular}{c|ccccc}
        \hline
        Dataset & Nodes & Edges & Classes & Features & homophily\\
        \hline
        Cora & 2708 & 5429 & 7 & 1433 & 0.83\\
        Citeseer & 3327 & 4732 & 6 & 3703 & 0.71\\
        Pubmed & 19717 & 44338 & 3 & 500 & 0.79\\
        \hline
        % Chameleon & 2277 & 36101 & 5 & 2325 & 0.25\\
        % Squirrel & 5201 & 217073 & 5 & 2089 & 0.22\\
        PT & 1912 & 31299 & 2 & 3169 & 0.59\\
        TW & 2772 & 63462 & 2 & 3169 & 0.55\\
        \hline
        Actor & 7600 & 33544 & 5 & 931 & 0.24\\
        \hline
        Cornell & 183 & 295 & 5 & 1703 & 0.30\\
        Texas & 183 & 309 & 5 & 1703 & 0.11\\
        Wisconsin & 251 & 499 & 5 & 1703 & 0.21\\
        \hline
        \end{tabular}
    }
% \vspace{-4.5mm}
\end{table}

\begin{itemize}[leftmargin=*]
    \item \textbf{Citation networks}~\cite{yang2016revisiting,welling2016semi}. Cora, Citeseer, and Pubmed are benchmark datasets of citation networks. Nodes represent paper, and edges represent citation relationships in these networks. The features are bag-of-words representations of papers, and labels denote their academic fields.
    % \item \textbf{Wikipedia networks}~\cite{rozemberczki2021multi,pei2019geom}. Wikipedia dataset contain three page to page networks, Chameleon, Squirrel and Crocodile, which are originally designed for website traffic regression. In ~\cite{pei2019geom}, the author package monthly traffic into 5 categories, providing Chameleon and Squirrel for node classification task. In both networks, vertices are web pages, edges are hyper-links and features are nouns that characterize the page.
    \item \textbf{Social networks}~\cite{rozemberczki2021multi}.
    TW and PT are two subsets of Twitch Gamers dataset~\cite{rozemberczki2021twitch}, designed for binary node classification tasks, where nodes correspond to users and links to mutual friendships. 
    The features are liked games, location, and streaming habits of the users. 
    The labels denote whether a streamer uses explicit language (Taiwanese and Portuguese).
    \item \textbf{WebKB networks}~\cite{getoor2005link}. 
    Cornell, Texas, and Wisconsin are three subsets of WebKB, where nodes are web pages, and edges are hyperlinks. The features are the bag-of-words representation of pages. The labels denote categories of pages, including student, project, course, staff, and faculty.
    % The WebKB dataset consists of 877 scientific publications classified into one of five classes. The citation network consists of 1608 links. Each publication in the dataset is described by a 0/1-valued word vector indicating the absence/presence of the corresponding word from the dictionary. The dictionary consists of 1703 unique words. 
    \item \textbf{Actor co-occurrence network}~\cite{tang2009social}. This dataset is a subgraph of the film-director-actor-writer network, in which nodes represent actors, edges represent co-occurrence relation, node features are keywords of the actor, and labels are the types of actors.
    % \item \textbf{Karateclub dataset}
\end{itemize}
% \vspace{-0.5em}

% \vspace{-1.0em}
\subsection{Baselines}
\label{appendix:baseline}
% Extensive baselines are used for comparison, which is briefly described as follows\footnote{For GCN, GAT, GraphSAGE, and APPNP layers, we adopt implementation from DGL library~\cite{wang2019deep}:https://github.com/dmlc/dgl }:
Baselines are briefly described as follows\footnote{For GCN, GAT, GraphSAGE, and APPNP layers, we adopt implementation from DGL library~\cite{wang2019deep}:https://github.com/dmlc/dgl }:
\begin{itemize}[leftmargin=*]
% \vspace{-1.5mm}
    \item \textbf{GCN}~\cite{welling2016semi} 
    is the most popular GNN, which defines the first-order approximation of a localized spectral filter on graphs.
    \item \textbf{GAT}~\cite{velivckovic2017graph}  
    introduces a self-attention mechanism to important scores for different neighbor nodes.
    \item \textbf{GraphSAGE}~\cite{hamilton2017inductive} 
    is an inductive framework that leverages node features to generate embeddings by sampling and aggregating features from the local neighborhood. 
    \item \textbf{APPNP}~\cite{gasteiger2019predict} 
    combines GCN with personalized PageRank. 
    \item \textbf{GCNII}\footnote{https://github.com/chennnM/GCNII}~\cite{chen2020simple}
    employs residual connection and identity mapping.
    \item \textbf{Grand}\footnote{https://github.com/THUDM/GRAND}~\cite{feng2020graph}
    purposes a random propagation strategy for data augmentation, and uses consistency regularization to optimize.
    \item \textbf{Mixhop}\footnote{https://github.com/samihaija/mixhop}~\cite{abu2019mixhop} aggregates mixing neighborhood information.
    \item \textbf{Geom-GCN}\footnote{https://github.com/graphdml-uiuc-jlu/geom-gcn}~\cite{pei2019geom}
    exploits geometric relationships to capture long-range dependencies within structural neighborhoods. Three variant of Geom-GCN is used for comparison.
    \item \textbf{GDC}\footnote{https://github.com/gasteigerjo/gdc}~\cite{gasteiger_diffusion_2019}
    refines graph structure based on diffusion kernels.
    % \item \textbf{Pro-GNN}\footnote{https://github.com/DSE-MSU/DeepRobust}~\cite{jin2020graph}
    \item \textbf{GEN}\footnote{https://github.com/BUPT-GAMMA/Graph-Structure-Estimation-Neural-Networks}~\cite{wang2021graph} 
    estimates underlying meaningful graph structures.
    \item \textbf{H$_2$GCN}\footnote{https://github.com/GemsLab/H2GCN}~\cite{zhu2020beyond} combine multi-hop neighbor-embeddings for adapting to both heterophily and homophily graph settings.
    \item \textbf{DropEdge}\footnote{https://github.com/DropEdge/DropEdge}~\cite{rong2019dropedge}
    randomly removes edges from the input graph for over-fitting prevention. 
    \item \textbf{Jaccard}\footnote{https://github.com/DSE-MSU/DeepRobust}~\cite{wu2019adversarial}
    prunes the edges connecting nodes with small Jaccard similarity.
\end{itemize} 

% \footnote{The implementation provided by CogDL~\cite{cen2021cogdl} is adopted for GCNII, GDC and Grand: https://github.com/thudm/cogdl}
% For SGC, GCN~\cite{welling2016semi}, Chebnet, GAT~\cite{velivckovic2017graph},GraphSAGE and APPNP, we adopt the implementations from the Deep Graph Learning library ~\cite{wang2019deep}.
% For the remaining baselines

% we utilize two-layer GNN encoders, and comparing with our \framework with them as the backbones. 

\begin{table*}[htb!]
    \renewcommand{\arraystretch}{0.95}
    \setlength{\abovecaptionskip}{0.25cm}
    \setlength{\belowcaptionskip}{-0.25cm}
    \caption{Comparison of training time(hr.) of achieving the best performance based on GPU.}
    % \vspace{-2.5mm}
    \label{tab:time comparasion}
    \centering
    % \scalebox{1.0}{
    \setlength{\tabcolsep}{3.6mm}{
        \begin{tabular}{l|ccccccccc}
        \hline
      Method & {Cora} & {Citeseer} & {Pubmed} & {PT} & {TW} & {Actor} & {Cornell} & {Texas} & {Wisconsin} \\
        \cline{1-1}
        \hline     % cora    cite    pub     pt    tw    act      cor      tex     wis
        \framework$_{GCN}$  &  0.071 & 0.213 & 4.574 & 0.178 &  0.374
                     & 1.482 & 0.006 & 0.008 & 0.009 \\
        \framework$_{SAGE}$   & 0.074 & 0.076 & 4.852 & 0.169 &  0.214
                     & 0.817 &  0.006 & 0.007 & 0.009 \\
        \framework$_{GAT}$   & 0.071 & 0.180 & 4.602 & 0.172 &  0.329
                     & 1.273 & 0.006 & 0.008 & 0.009 \\
        \framework$_{APPNP}$   & 0.073 & 0.215 & 4.854 & 0.138 &  0.379
                     & 1.367 & 0.010 & 0.011 & 0.013 \\
        \hline
        \end{tabular}
    }
    % \vspace{-2.5mm}
\end{table*}

% \vspace{-1.0em}
\subsection{Overall algorithm of \framework}
The overall algorithm of \framework~ is shown in Algorithm ~\ref{algorithm:training}. Note that, if choose to retain the connection from the previous iteration, to ensure that the number of edges remains stable during the training, a percentage of edges in the reconstructed graph with low similarity will be dropped in each iteration.

\label{appendix:overall algorithm}
\begin{algorithm}[htb!]
\SetAlgoRefName{1}
\SetAlgoVlined
\KwIn{a graph $G=(V,E)$, features $X$, labels $Y_L$, mode $\in {True,False}$\\
iterations $\eta$, encoding tree height $K$, hyperparameter $\theta$}
\KwOut{optimized graph $G'=(V,E')$, prediction $Y_P$, GNN parameters $\Theta$ }
Initialize $\Theta$;\\
\For{$i=1$ to $\eta$}{
    Update $\Theta$ by classification loss $\mathcal{L}_{cls}(Y_L,Y_P)$;\\
    Getting node representation $X'=\mathrm{GNN}(X)$;\\
    Initialize $k=1$ for $k$-NN structuralization;\\
    Create fusion map $G_{f}$ according to Algorithm~\ref{algorithm:kselector};\\
    Create $K$-dimensional encoding tree $\mathcal{T}^*$ according to Algorithm~\ref{algorithm:KDimSEMinimize};\\
    \For{each non-root node $\alpha$ in $\mathcal{T}^*$}{
        Calculate $H^{\mathcal{T}^*}(G_{f};(\lambda,\alpha])$ through Eq.~\ref{eq:deduct-se};\\
        Assign probability $P(\alpha)$ to $\alpha$ through Eq.~\ref{eq:prob-softmax};\\
    }
    \For{each subtree rooted at $\alpha$ in $\mathcal{T}^*$}{
        Assuming $\alpha$ has $n$ children, set $t = \theta \times n$;\\ 
        \For{$j=1$ to $t$}{
            Sample a node pair $(v_m,v_n)$ according to \S~\ref{step3};\\
            Adding edge $e_{mn}$ to $G'$;\\
        }
    }
    \If{mode}{
        Let $E' = E \cup E'$, where $E'$ and $E$ are the edge set of $G'$ and $G$, respectively;\\
        Drop a percentage of edges in $G'$;
    }
    Update graph structure $G \gets G'$;
    Update node representation: $X \gets X'$;\\
}
Get prediction $Y_P$;\\
Return $G'$, $Y_P$ and $\Theta$;\\
\caption{Model training for~\framework}
\label{algorithm:training}
\end{algorithm}

% \vspace{-1.0em}
\subsection{Algorithm of one-dimensional structural entropy guided graph enhancement}
\label{appendix:1dse algorithm}
The $k$-selector is designed for choosing an optimal $k$ for $k$-NN structuralization under the guidance of one-dimensional structural entropy maximization. The algorithm of $k$-selector and fusion graph construction is shown in Algorithm~\ref{algorithm:kselector}.

\begin{algorithm}[htb!]
\SetAlgoRefName{2}
\SetAlgoVlined
\KwIn{a graph $G=(V,E)$, node representation $X$}
\KwOut{fusion graph $G_{f}$ }
Calculate $S \in \mathbb{R}^{|V|\times |V|}$ via Eq.~\ref{eq:pcc};\\
\For{$k=2$ to $|V|-1$}{
    Generate $G_{knn}$ by $S$;\\
    Generate $G^{(k)}_{f} = \{V,E_{f} = E \cup E_{knn} \}$;\\
    Reweight $G^{(k)}_{f}$ via Eq.~\ref{eq:reweighted};\\
    Calculate $H^1(G^{(k)}_{f})$ via Eq.~\ref{eq:H1};\\
    \If{$H^1(G^{(k)}_{f})$ reaches the maximal optima}{
        $G_{f} \gets G^{(k)}_{f}$;\\
        Return $G_{f}$;
    }
}
\caption{$k$-selector and fusion graph construction}
\label{algorithm:kselector}
\end{algorithm}

% \vspace{-1.0em}
\subsection{Algorithm of high-dimensional structural entropy minimization}
\label{appendix:kdse algorithm}
% \vspace{-0.5em}
% The high-dimensional structural entropy minimization is a heuristic algorithm that adjusts encoding tree structure by greedily executing specific operators. The pseudo-code is shown in Algorithm~\ref{algorithm:KDimSEMinimize}.
The pseudo-code of the high-dimensional structural entropy minimization algorithm is shown in Algorithm~\ref{algorithm:KDimSEMinimize}.
%%伪代码
\begin{algorithm}[htb!]
\SetAlgoRefName{3}
\SetAlgoVlined
\KwIn{a graph $G=(V,E)$, the height of encoding tree $k>1$}
\KwOut{Optimal high-dimensional encoding tree $\mathcal{T}^*$}
//Initialize an encoding tree $\mathcal{T}$ with height $1$ and root $\lambda$\\
Create root node $\lambda$;\\
\For{$v_i \in V$}{
    Create node $\alpha_i$. Let $T_{\alpha_i} = v_i$;\\
    $v_i^- = \lambda$;\\
}
//Generation of binary encoding tree\\
%% problem
\While{$\lambda$ has more than $2$ children}{
    Select $\alpha_i$ and $\alpha_j$ in $\mathcal{T}$, 
    condition on $\alpha_i^-=\alpha_j^-=\lambda$ and $\mathop{\arg\max}\limits_{\alpha_i,\alpha_j}(H^\mathcal{T}(G)-H^\mathcal{T}_{\mathrm{CB}(\alpha_{i},\alpha_{j})}(G))$;\\
    $ \mathrm{CB}(\alpha_{i},\alpha_{j})$ according to Definition ~\ref{def:CBop};\\
}
//Squeezing of encoding tree\\
\While{$\mathrm{height}(\mathcal{T}) > K$}{
    Select non-root node $\alpha$ and $\beta$ in $\mathcal{T}$, 
    condition on $\alpha^-=\beta$ and $\mathop{\arg\max}\limits_{\alpha,\beta}(H^\mathcal{T}(G)-H^\mathcal{T}_{\mathrm{LF}(\alpha,\beta)}(G))$;\\
    $ \mathrm{LF}(\alpha,\beta)$ according to Definition ~\ref{def:LFop};\\
}
Return $\mathcal{T}^* \gets \mathcal{T}$;
\caption{K-dimensional structural entropy minimization}
\label{algorithm:KDimSEMinimize}
\end{algorithm}

\end{document}